\documentclass[titlepage]{csetr}

\usepackage{graphicx}
\usepackage{amsmath,amsfonts,amsthm,amssymb}
\usepackage{setspace}
\usepackage{url}
\usepackage{fancyhdr}
\usepackage{lastpage}
\usepackage{multicol}
\usepackage{pdfpages}
\usepackage{color}
\usepackage{xcolor}
\usepackage[normalem]{ulem}
\usepackage{booktabs}
\usepackage{multirow}
\usepackage{algorithm}
\usepackage[noend]{algpseudocode}
\usepackage{times}
\usepackage{helvet}
\usepackage{courier}
\usepackage{latexsym}
\usepackage[noend]{algpseudocode}
\usepackage{subcaption}
\usepackage{amssymb}
\usepackage{pifont}
\usepackage{lipsum}
\usepackage{capt-of}

\renewcommand{\and}{\hspace{.5cm}}

\trnumber{201706}
\trdate{February 2017}

\title{%
  A Semantically Motivated Approach to Compute \textsc{Rouge} Scores
}
\author{%
  Elaheh ShafieiBavani \and %
  Mohammad Ebrahimi \and %
  Raymond Wong\\[1em]
  Fang Chen\\[2em]
  University of New South Wales, Australia \\%
  Data61 CSIRO, Sydney, Australia \\%
  \email{\{elahehs,mohammade,wong,fang\}@cse.unsw.edu.au}\\[3cm]
}

\date{}
\begin{document}
\maketitle

\begin{abstract}
	\begin{quote}
		\textsc{Rouge} is one of the first and most widely used evaluation metrics for text summarization. However, its assessment merely relies on surface similarities between peer and model summaries. Consequently, \textsc{Rouge} is unable to fairly evaluate abstractive summaries including lexical variations and paraphrasing. Exploring the effectiveness of lexical resource-based models to address this issue, we adopt a graph-based algorithm into \textsc{Rouge} to capture the semantic similarities between peer and model summaries. Our semantically motivated approach computes \textsc{Rouge} scores based on both lexical and semantic similarities. Experiment results over TAC AESOP datasets indicate that exploiting the lexico-semantic similarity of the words used in summaries would significantly help \textsc{Rouge} correlate better with human judgments.
	\end{quote}
\end{abstract}

\section{Introduction} 

Quantifying the quality of summaries is an important and necessary task in the field of automatic text summarization. Traditionally, this task involves a human assessment of various quality criteria (e.g. coherence, conciseness, grammaticality, informativity and readability) \cite{mani2001automatic}. Therefore, manual evaluation requires a lot of time and expertise in the field of given texts. To tackle this issue, automatic evaluation metrics come into play. This advent opens a new door to meta-evaluation (i.e. evaluation of evaluation metrics \cite{ellouze2013evaluation}). On the importance of meta-evaluation and its impact on summarization research, Text Analysis Conference (TAC\footnote{http://www.nist.gov/tac}) provides the task of Automatically Evaluating Summaries of Peers (AESOP) to assess the correlation of evaluation metrics with human judgments.      
\newline \indent Among the proposals for automatic evaluation metrics \cite{hovy2006automated,tratz2008bewte,giannakopoulos2008summarization}, \textsc{Rouge}\footnote{Recall-Oriented Understudy for Gisting Evaluation} \cite{lin2004rouge} is the first and still most widely used one  \cite{graham2015re}. This metric measures the concordance of system-generated summaries (peer summaries) and human-generated reference summaries (model summaries) by determining n-grams, word sequences, and word pair matches. \textsc{Rouge} has frequently been proven to correlate very well with human judgments \cite{lin2004looking,owczarzak2011overview,over2004introduction}. However, its assessment heavily relies on surface similarities between peer and model summaries. Hence, it is unable to fairly evaluate abstractive summaries which might include semantically similar units with different lexical representations (e.g. paraphrasing).
\newline \indent For more clarity, consider the following two sentences: \textit{(i) They strolled around the city; (ii) They took a walk to explore the town}. These sentences are semantically the same, but lexically different. If one of them is included in a model summary, while a peer summary contains another one, \textsc{Rouge} or other surface based evaluation metrics cannot capture their similarity due to the minimal lexical overlap. Our aim is to help \textsc{Rouge} with identifying the semantic similarities of linguistic items at the deepest sense level, and consequently tackling the main problem of its bias towards lexical similarities. 
\newline \indent Considering senses instead of words, we make use of the Personalized PageRank (PPR) algorithm \cite{haveliwala2002topic} to leverage repetitive random walks on WordNet 3.0 \cite{fellbaum1998wordnet} as a semantic network, and obtain the probability distribution of each disambiguated sense over all senses in the network. The weights in this distribution denote the relevance of the corresponding senses. Our graph-based approach (\textsc{GRouge}) favors semantic similarity scores between n-grams, along with their match counts (used originally in \textsc{Rouge}), to perform both semantic and lexical comparisons of a peer summary text and a set of model summaries.
\newline \indent To demonstrate the effectiveness of our approach, we have conducted a set of experiments over the TAC 2010 and 2011 AESOP datasets. We have compared the output of \textsc{GRouge} with three manual metrics of Pyramid, Readability, and Responsiveness. The results we have achieved via three metrics of correlation (i.e. Pearson, Spearman, Kendall) demonstrate that \textsc{GRouge} variants significantly outperform their corresponding variants of \textsc{Rouge} most of the time. Beyond just enhancing the evaluation prowess of \textsc{Rouge}, this approach has the potential to expand the applicability of \textsc{Rouge} to abstractive summarization as well. The rest of the paper is organized as follows. Section \ref{sec:backg} summarizes the background. The proposed approach is explained in Section \ref{sec:proposed}. Section \ref{sec:expr} reports the utilized data, the performed meta-evaluation, and the achieved results. Finally, Section \ref{sec:concl} concludes the paper.

\section{Background}\label{sec:backg}
\textsc{Rouge} includes a large number of distinct variants, including four methods of n-gram counting (\textsc{Rouge-N; S; W; L}). In summarization literature, a few of these variants are often chosen arbitrarily to assess the quality of summarization approaches. \textsc{Rouge-1}, \textsc{Rouge-2}, and \textsc{Rouge-su4} are reported to have a strong correlation with human assessments, and are frequently used to evaluate summaries \cite{lin2004looking,owczarzak2011overview,over2004introduction}. \textsc{Rouge-1} and 2, respectively calculate unigram and bigram co-occurrence statistics. \textsc{Rouge-su4} measures co-occurring bigrams with maximum skip distance 4. It is noteworthy that \textsc{Rouge-2} and \textsc{su4} have been defined as baseline systems in TAC AESOP task. 

Although \textsc{Rouge} is a popular evaluation metric, studies on improving the current evaluation metrics is still an open research area. Many of these efforts are analyzed and gathered in a survey provided by \cite{steinberger2012evaluation}. In this section, we try to briefly review the most significant ones. Since DUC 2005, the Pyramid metric \cite{passonneau2005applying} was introduced as one of the principal metrics for evaluating summaries in the TAC conference. However, this metric is semi-automated and requires manual identification of summary content units (SCUs). The approach proposed in \cite{hovy2006automated} is based on comparison of basic syntactic units, so called Basic Elements (BE) between the peer and model summaries. This metric, namely BE-HM was specified as one of the baselines in the TAC AESOP task. Among participating systems in this task from 2009 to 2011, AutoSummENG \cite{giannakopoulos2008summarization} was reported as one of the top systems. This graph-based metric (\textsc{DemokritosGR} in the experiments), compares the graph representations of peer and model summaries. 

Surface-based evaluation metrics work well as long as a surface-based summary (i.e. extractive) is to be assessed. Difficulties arise while evaluating abstractive summaries including terminology variations and paraphrasing. For example, consider the following two phrases \cite{ng2015better}: \textit{(i) It is raining heavily; (ii) It is pouring}. If we are performing a lexical string match, as \textsc{Rouge} does, there is nothing in common between the terms "raining", "heavily", and "pouring". However, these two phrases are semantically the same. Hence, we study the effectiveness of semantically motivated approaches to measure word semantic similarity on improving \textsc{Rouge} evaluation. For this purpose, approaches can be grouped into two categories of distributional and lexical resource-based \cite{pilehvar2015senses}. A recent branch of distributional models uses neural networks to directly learn the expected context of a given word and model it as a continuous vector \cite{turian2010word,baroni2014don}, often referred to as word embedding. In the context of summarization evaluation, an automated variant of the Pyramid metric which uses word embeddings to map text content within peer summaries to SCUs has recently been proposed \cite{passonneau2013automated}. However, the SCUs still need to be manually identified. To overcome this deficiency, a more recent automatic metric \cite{ng2015better}, namely \textsc{Rouge-WE} has enhanced \textsc{Rouge} by incorporating the use of a variant of word embeddings, called word2vec \cite{mikolov2013linguistic}. However, a good performance for Word2vec is usually obtained upon tuning different configurations of this model on a large number of different datasets \cite{baroni2014don}. 

Lexical resource-based approaches usually make an assumption that the similarity of two words can be calculated in terms of the similarity of their closest senses. Among them, a random walk-based method that models disambiguated words through the distributions of the PPR algorithm on the WordNet graph has proven to be promising \cite{pilehvar2015senses}. Unlike this approach, none of the above-mentioned techniques disambiguates the words being compared, and they hence consider a word as a conflation of all its meanings, which potentially reduces the quality of similarity measurement. Therefore, we are prompted to disambiguate n-gram pairs to a set of intended senses prior to modeling. This will make us able to identify the semantic similarity of peer and model summaries, independently of their surface forms or any semantic ambiguity therein. 

Given a pair of peer and model summaries, we first utilize the PPR algorithm to acquire probability distributions of their words' senses over the WordNet graph (PPR vectors). Comparing these vectors obtained for all senses in a pair of peer and model summaries, we disambiguate each word into its most appropriate sense. This helps us to measure the semantic similarity of n-grams at the deepest sense level. PPR vector is calculated this time for each of the model-gram (each n-gram in the model summary) and the peer summary text by initializing random walks from their disambiguated senses over WordNet. We further compute their semantic similarity by comparing the resulting PPR vectors. This remedy is finally adopted into \textsc{Rouge} variants, and the appropriate weights of lexical and semantic similarity scores are explored through our experiments.

\section{The Proposed Approach}\label{sec:proposed}
\textsc{Rouge} assumes that a peer summary is of high quality if it shares many words or phrases with a model summary. However, different terminology may be used to refer to the same concepts and hence relying only on lexical overlaps may underrate content quality scores. To tackle this issue, our approach utilizes both semantic and lexical similarities between a peer and its corresponding model summary. This method also enables us to reward terms that are not lexically equivalent, but semantically related. 

\subsection{Measuring Semantic Similarities}\label{sec:semsim}            
Given a pair of peer and model summaries, we need to compute and compare PPR vectors at the following levels: \textit{(i) sense level}, to disambiguate each word (having a set of senses); and \textit{(ii) n-gram level}, to measure the semantic similarity. Next, we explain how a PPR vector is constructed for a sense or a set of senses, and how a similarity score is computed accordingly.

\paragraph{PPR Vectors:} To construct a PPR vector, we perform iterative random walks beginning at a sense (seed) or a set of senses (seeds) on WordNet. This provides a frequency or multinomial distribution over all senses in WordNet \cite{pilehvar2013align}. A higher probability will be assigned to senses that are frequently visited from the seeds. This representation is applicable both when the item itself is a single sense and when the item is a sense-tagged text. For better clarity, consider an adjacency matrix $W$ for the WordNet graph, where edges connect senses according to the relations defined in WordNet (e.g. hypernymy, synonymy, etc.). The probability distribution for the starting location of the random walker in the network is denoted by $\vec{V}^{\,(0)}$. Given the set of senses $S$ in a lexical item, the probability mass of $\vec{V}^{\,(0)}$ is uniformly distributed across the senses $s_i \in S$, with the mass for all $s_i \notin S$ set to zero. The PPR vector is then computed using Equation \ref{eq:ppr}.

\begin{equation}
\vec{V}^{\,(t)}=(1-\alpha)W\vec{V}^{\,(t-1)}+\alpha\vec{V}^{\,(0)}
\label{eq:ppr}
\end{equation}

\noindent where at each iteration, the random walker may jump to any node $s_i \in S$ with probability $\alpha/|S|$. Following the standard convention, the value of $\alpha$ is set to 0.15. The number of iterations is also set to 30, which is sufficient for the distribution to converge. The resulting probability vector $\vec{V}^{\,(t)}$ is the PPR vector of the lexical item, as it has aggregated its senses' similarities over the entire graph. The UKB\footnote{http://ixa2.si.ehu.es/ukb/} implementation of PPR is used to this end. 

To compare the PPR vectors of each pair of n-grams, we use an effective method, namely Weighted Overlap, which has consistently proven to be superior to cosine similarity, Jensen-Shannon divergence, and Rank-Biased Overlap for comparing vectors in different datasets \cite{pilehvar2015senses}. This algorithm first sorts the two vectors according to their values and then harmonically weights the overlaps between them. Finally, the semantic similarity ($Sim_{sem}$) of two vectors $V_1$ and $V_2$ is calculated by Equation \ref{eq:semsim}.

\begin{equation}
Sim_{sem}(V_1,V_2)=
\begin{cases}
\frac{\sum_{h \in H}{(r_h(V_1)+r_h(V_2))^{-1}}}{\sum_{i=1}^{|H|}{(2i)^{-1}}}, & \text{if}\ |H|>0 \\
0, & \text{otherwise}
\end{cases}
\label{eq:semsim}
\end{equation}

\noindent where \textit{H} denotes the intersection of all senses with non-zero probability (dimension) in both vectors, and $r_h(V_j)$ denotes the rank of the dimension \textit{h} in the sorted vector ${V_j}$, where rank 1 denotes the highest rank. The denominator is also used as a normalization factor that guarantees a maximum value of one. The minimum value is zero and occurs when there is no overlap between the two vectors, i.e. $|H|=0$. Next, we explain the process of n-gram disambiguation into a set of appropriate senses.

\paragraph{Disambiguation of n-grams:} Prior to measuring semantic similarities, each word in the n-grams has to be analyzed and disambiguated into its intended sense. However, conventional word sense disambiguations are not applicable due to the lack of contextual information. Hence, we make use of an alignment-based algorithm proposed by \cite{pilehvar2013align} to disambiguate each word. This algorithm seeks the semantic alignment that maximizes the similarity of the senses of the compared words. In our approach, given two n-grams, for each word type $t_i$ in n-gram $G_1$, this algorithm assigns $t_i$ to the sense that has the maximal similarity score to any sense of the word types in the compared n-gram $G_2$. As an example, let us consider two sentences of {\it "a1. Officers fired."} and {\it "a2. Several policemen terminated in corruption probe."}, the semantic alignment procedure has been performed as follows \cite{pilehvar2015senses}: \vspace{0.3cm} 

$P_{a1}.$ {\it officer}$_{n}^{3}, fire_{v}^{4}$

$P_{a2}.$ {$policeman_n^1, terminate_v^4, corruption_n^6, probe_n^1$}\vspace{0.3cm}

\noindent where $P_j$ denotes the corresponding set of senses of sentence \textit{j}. $t_p^i$ denotes the $i$-th sense of a word $t$ in WordNet with part-of-speech $p$. After alignment, among all possible pairings of all senses of $fire_v$ to all senses of all words in \textit{a2}, the sense $fire_v^4$ (employment termination) obtains the value $Sim_{sem}(fire_v^4,terminate_v^4)=1$, which is the maximal similarity value. $Sim_{sem}$ gives the semantic similarity of two senses by comparing their PPR vectors, as defined in Equation \ref{eq:semsim}. Therefore, \textsc{GRouge} transforms the task of determining overlapping n-grams in \textsc{Rouge} into that of computing the similarity of the best-matching sense pair across the two n-grams. It also enables the same n-grams to have different meanings when paired with different linguistic items. In the following, the generated PPR vectors for a pair of disambiguated model-gram and peer summary are compared to calculate their semantic similarities.

\begin{figure}[h!]
	\centering
	\includegraphics[scale=0.20]{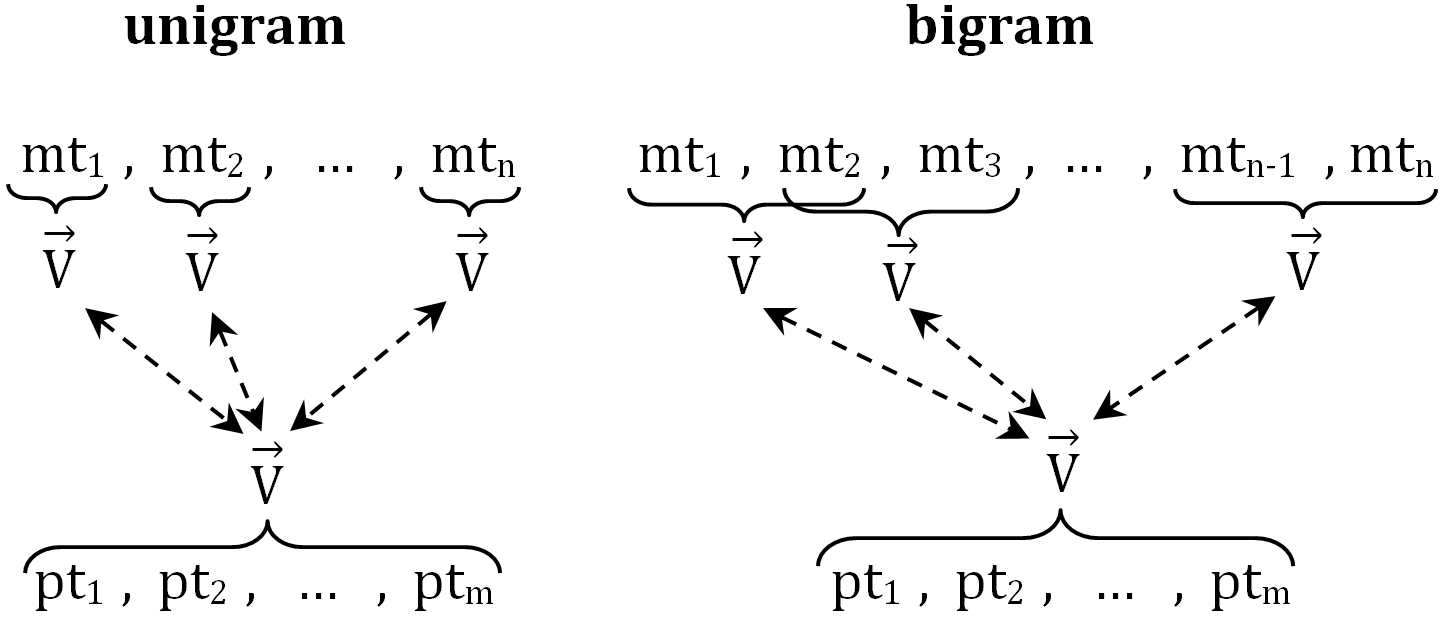}
	\caption{Comparing PPR vectors between model-grams (unigrams and bigrams in a model summary text) and a peer summary text}
	\label{fig:gram}
\end{figure}

\paragraph{Model-grams against Peer Summary:} Exploiting underlying semantic similarities between all n-gram pairs in the model and peer summary texts takes a lot of time and effort. To overcome this issue, we consider the peer summary text as a sense-tagged unit, and measure its semantic similarity against each n-gram in the model summary text (model-gram). For better clarity, let us consider $MT=\{mt_1, mt_2, ..., mt_n\}$, and $PT=\{pt_1, pt_2, ..., pt_m\}$ as the sets of tokens of a model and a peer summary text, respectively. Figure \ref{fig:gram} shows how PPR vectors of unigrams and bigrams in a model summary text are compared to the PPR vector of the peer summary text. 

Measuring semantic similarities and sense disambiguation are previously explained in details. We can list the steps as follows: (i) Generating PPR vectors for all senses in the model-gram and peer summary text; (ii) Comparing the PPR vectors to disambiguate the model-gram and peer summary text to a set of proper senses; (iii) Generating one PPR vector for each of the model-gram and peer summary text by initializing random walks from their disambiguated senses over WordNet; (iv) Comparing the resulting PPR vectors to compute the semantic similarity between the model-gram and peer summary text. Treating the peer summary text as one unit not only reduces comparison time and increases the efficiency, but also provides a suitable number of content words which guarantees implicit word sense disambiguation, and semantic relationship derivation.  

\subsection{OOV Handling}

Similarly to any other graph-based approach that maps words in a given textual item to their corresponding nodes in a semantic network, modeling n-grams through PPR vectors can suffer from the limited coverage of words. This means that only those words that are associated with some nodes in WordNet can be handled. Since out-of-vocabulary (OOV) words are the words that are not defined in the corresponding lexical resource, they will be ignored while generating PPR vectors. The reason is that they do not have an associated node in the WordNet graph for the random walk to be initialized from. Denying OOV words, such as infrequent named entities, acronyms or jargon, while increasing in a text, can be problematic when measuring semantic similarity of n-gram pairs. To take OOV words into consideration, we follow the approach proposed by \cite{pilehvar2015senses} and directly insert each OOV word into the resulting PPR vector. To this end, we introduce new dimensions in the resulting PPR vector, one for each OOV term, while assigning a weight to the new dimension so as to guarantee its placement among the top dimensions in its PPR vector.

\subsection{Multiple Levels of Evaluation}   
Most single automatic metrics use one level of evaluation (i.e. lexical, syntactic or semantic). A better approach is to assess the results while combining multiple levels of evaluation into one model \cite{ellouze2013evaluation}. For better clarity, consider the following groups of sentences:\newline\newline
\textit{a1. Soldiers were killed.}\newline
\textit{a2. Soldiers were executed.}\newline
\textit{a3. Military personnel were executed for committed crimes.}\newline\newline
\textit{b1. Soldiers were killed.}\newline
\textit{b2. Soldiers were murdered.}\newline
\textit{b3. Several servicemen were murdered by criminals.}\newline

Surface-based approaches that are merely based on string similarity cannot capture the similarity between any of the above pairs of \textit{a1} and \textit{a3}, or \textit{b1} and \textit{b3} as there exists no lexical overlap. In addition, a surface-based semantic similarity approach considers both \textit{a1} and \textit{b1} as being identical sentences, whereas we know that different meanings of the verb "kill" are triggered in the two contexts. Although verbs "kill", "execute" and "murder" are close together in WordNet, \textit{a2} and \textit{b2} carry very different connotations. As a remedy, we need to transform words to senses and perform disambiguation by taking into account the context of the paired linguistic item, hence providing a deeper measure of similarity comparison. We finally combine the lexical and semantic similarity scores to calculate \textsc{GRouge-N} (Equation \ref{eq:grouge}). This approach can increase the chance of getting the evaluation results more correlated with human assessments. 

\begin{equation}
\textsc{GRouge-N}=\frac{\displaystyle\sum_{S\in\{ModelSummaries\}}{\displaystyle\sum_{gram_n\in S}{Sim_{LS}({gram_n},P)}}}{\displaystyle\sum_{S\in\{ModelSummaries\}}{\displaystyle\sum_{gram_n\in S}{Count({gram_n})}}}
\label{eq:grouge}
\end{equation} 

\noindent where $n$ stands for the length of n-gram, and $Sim_{LS}$ is the score of lexico-semantic similarity between a model-gram $(gram_n)$ and the peer summary text $(P)$. 

To compute $Sim_{LS}$, we have conducted a set of experiments using lexical similarities, $Count_{match}(gram_n, P)$, and/or semantic similarities, $Sim_{sem}$ (Equation \ref{eq:semsim}). Note that $Count_{match}(gram_n, P)$ is the maximum number of n-grams co-occurring in a peer summary and a set of model summaries. The best correlation is obtained while using a linear combination of both scores with different weights according to Equation \ref{eq:finalsim}. 

\begin{equation}
Sim_{LS}(gram_n, P)=\beta \times Count_{match}(gram_n, P) + (1 - \beta) \times Sim_{sem}(gram_n, P)
\label{eq:finalsim}
\end{equation} 

The scaling factor $\beta$ was optimized on the TAC 2010 AESOP dataset \cite{owczarzak2010overview}, and set to 0.5 to reach the best correlation with the manual metrics of Pyramid and Responsiveness.

\section{Experiments}\label{sec:expr}

\subsection{Data and Meta-evaluation}

For the task of summarization evaluation, TAC has provided two benchmark AESOP datasets (AESOP 2010 and 2011), on which we can assess \textsc{GRouge}. We make use of the TAC 2010 AESOP dataset to optimize the scaling factor, and the TAC 2011 AESOP dataset to evaluate \textsc{GRouge}. This dataset consists of 44 topics, and two sets of 10 documents for each topic: set A for initial summaries; set B for update summaries. There are four human-crafted model summaries for each document set. A summary for each topic is generated by each of the 51 summarizers which participated in the main TAC summarization task. Source documents for summarization are taken from the New York Times, the Associated Press, and the Xinhua News Agency newswire. 

Two different types of evaluation were tasked in TAC 2011 AESOP: \textit{All Peers} and \textit{No Models}. The former case assigns a score to each peer summary, including the model summaries. This evaluation is intended to focus on whether an automatic metric can distinguish between human and automatic summarizers. The latter assigns a score to each peer summary, excluding the model summaries. This case is intended to focus on how well an automatic metric is able to assess automatic summaries. Using model summaries as references, each automatic summary can be evaluated against all four references simultaneously. Since our aim is to evaluate the quality of automatic summaries, we make use of \textit{No Models} evaluation.

The output of participating automatic metrics is tasked to be compared against human judges using three manual metrics of \textit{Pyramid}, \textit{Readability}, and \textit{Responsiveness}. Hence, the outputs are scored based on their summary content, linguistic quality, and a combination of both, respectively. Prior to computing correlation of \textsc{GRouge} variants with manual metrics, \textsc{GRouge} scores have reliably been computed (95\% confidence intervals) under \textsc{Rouge} bootstrap resampling with the default number of sampling point = 1000. Correlation of \textsc{GRouge} evaluation scores with the human judgments is then assessed with three metrics of correlation: Pearson $r$; Spearman $\rho$; and Kendall $\tau$.  

The value of all measures is between -1 and 1 of which 1 or -1 indicates a strong relationship between the two measures. The closer the value is to zero, the weaker the relation between the two measures. 25 automatic metrics participated in the TAC 2011 AESOP task, three of which (i.e. \textsc{Rouge-2}, \textsc{Rouge-su4}, and BE-HM) were used as baselines. In our experiments, the effectiveness of \textsc{GRouge} is demonstrated by assessing its three variants (\textsc{GRouge-1}, 2, and \textsc{su4}) against their corresponding variants of \textsc{Rouge}, and the other 23 AESOP participants. Note that \textsc{Rouge-1} was not among the participating metrics, but will be considered in our experiments. We compute scores using the default NIST settings for baselines in the TAC 2011 AESOP task (with stemming and keeping stopwords\footnote{https://tac.nist.gov/2011/Summarization/AESOP.2011.
	guidelines.html}). 

\subsection{Results}

\begin{figure}[h!]
	\centering
	\includegraphics[scale=0.65]{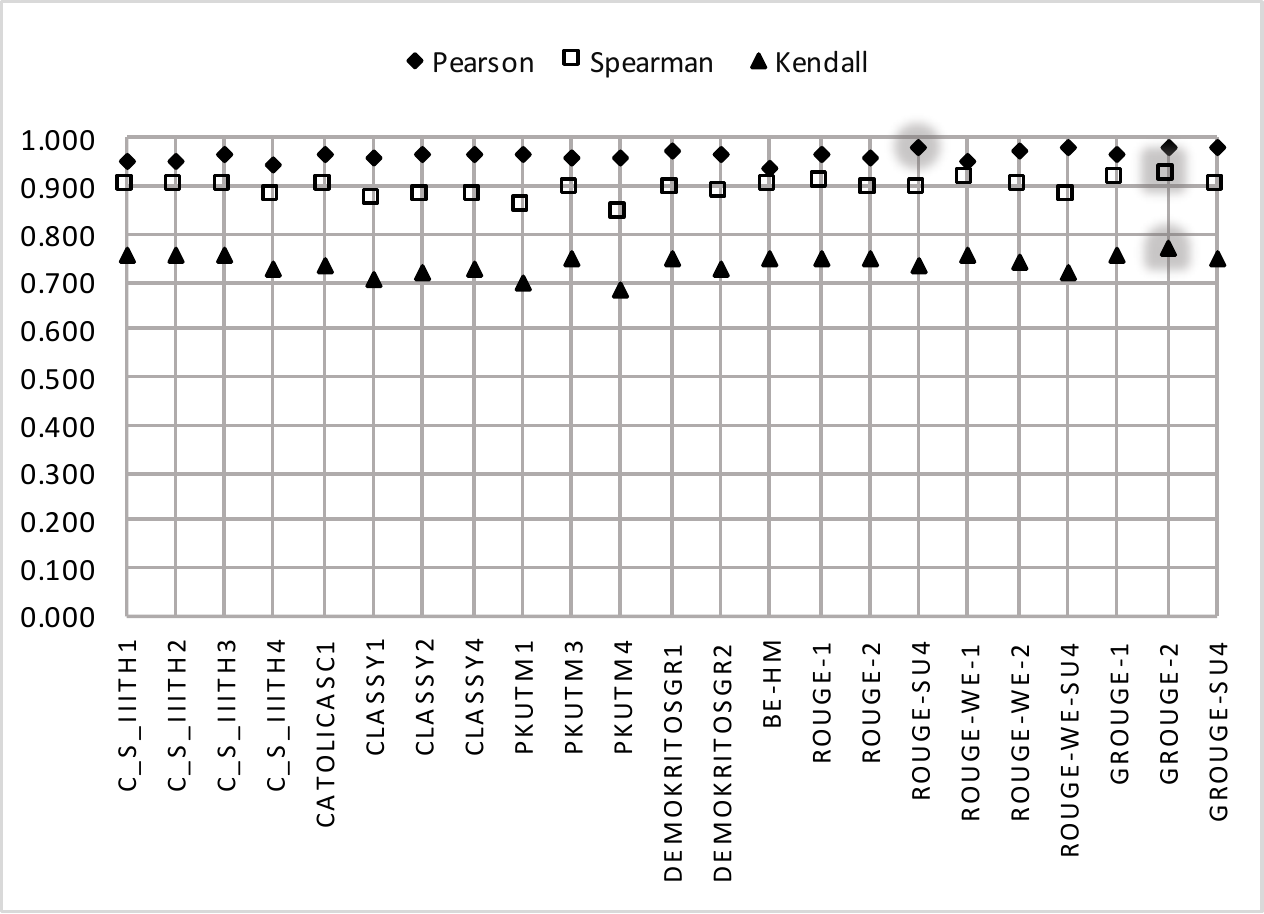}
	\caption{Correlation of the \textbf{best-performing AESOP metrics} with the manual metric of Pyramid using the correlation metrics of Pearson $r$, Spearman $\rho$, and Kendall $\tau$ on the TAC 2011 AESOP dataset}
	\label{fig:pyr}
\end{figure}

\begin{figure}[h!]
	\centering
	\includegraphics[scale=0.65]{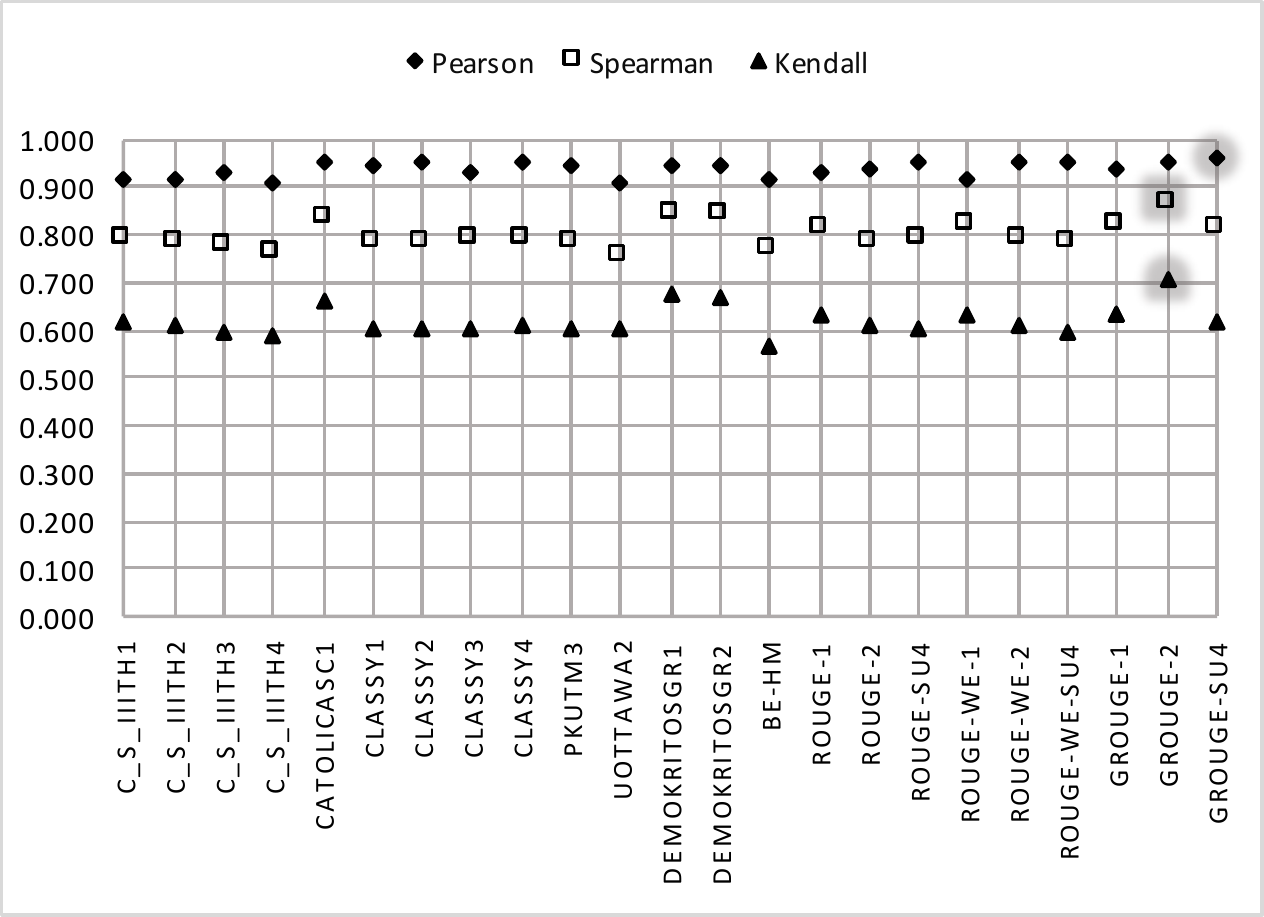}
	\caption{Correlation of the \textbf{best-performing AESOP metrics} with the manual metric of Responsiveness using the correlation metrics of Pearson $r$, Spearman $\rho$, and Kendall $\tau$ on the TAC 2011 AESOP dataset}
	\label{fig:resp}
\end{figure}

\begin{figure}[h!]
	\centering
	\includegraphics[scale=0.65]{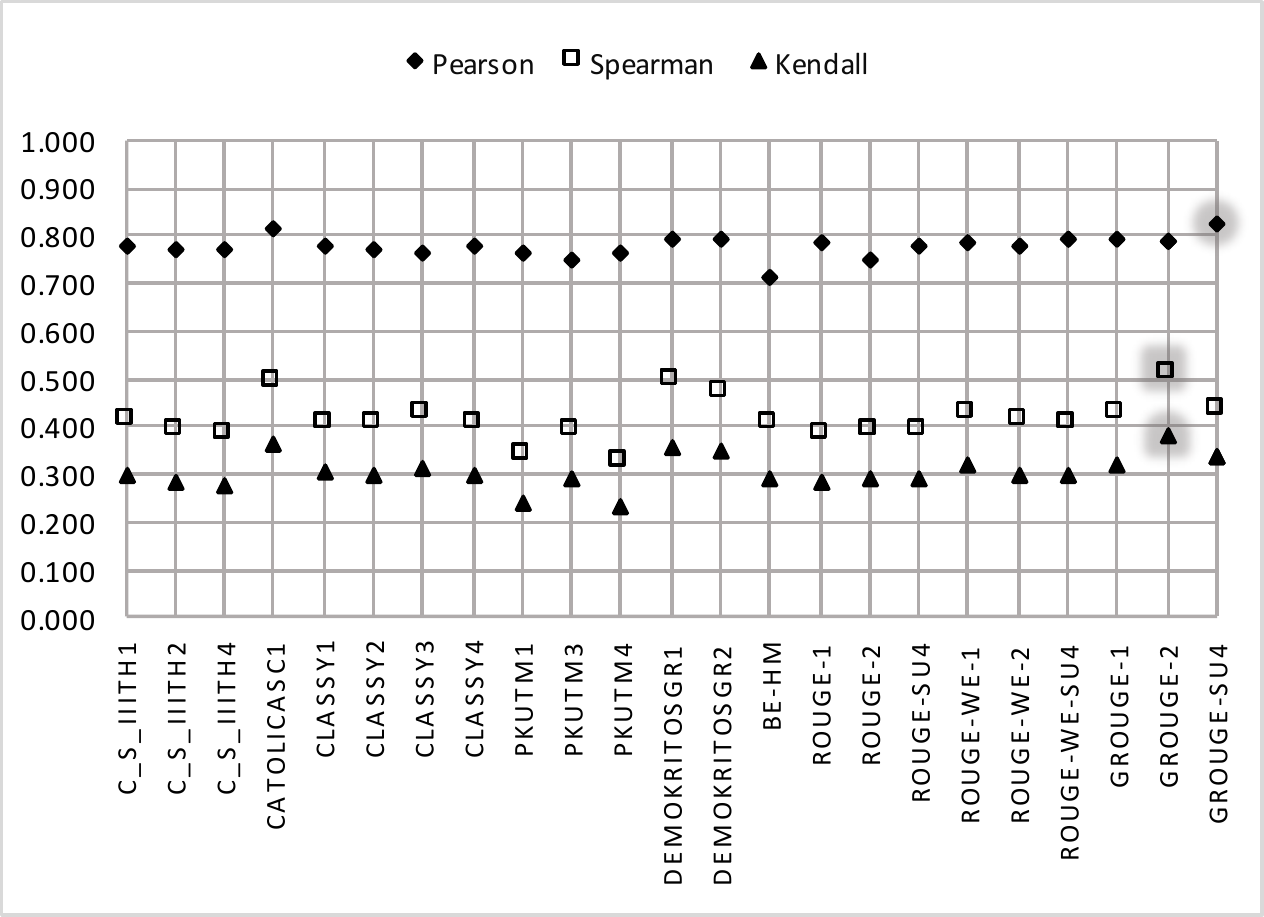}
	\caption{Correlation of the \textbf{best-performing AESOP metrics} with the manual metric of Readability using the correlation metrics of Pearson $r$, Spearman $\rho$, and Kendall $\tau$ on the TAC 2011 AESOP dataset}
	\label{fig:read}
\end{figure}

We have conducted a set of experiments to evaluate three variants of \textsc{GRouge} (i.e. \textsc{GRouge-1}, 2, and \textsc{su4}), against the top 13 best-performing metrics among the 22 metrics participated in AESOP, the baselines (i.e. \textsc{Rouge-2}, \textsc{su4}, BE-HM), \textsc{Rouge-1}, and the most recent related work (\textsc{Rouge-WE}). Correlation results of the best-performing AESOP metrics with Pyramid, Responsiveness, and Readability scores to the correlation metrics of Pearson $r$, Spearman $\rho$, and Kendall $\tau$ are depicted in Figures \ref{fig:pyr}, \ref{fig:resp}, and \ref{fig:read}, respectively. The highest correlation results are highlighted for better clarity. To demonstrate the effectiveness of \textsc{GRouge} in the \textsc{Rouge} framework, the obtained correlation results of all variants of \textsc{Rouge}-based metrics (\textsc{Rouge}, \textsc{Rouge-WE}, and \textsc{GRouge}) with Pyramid, Responsiveness, and Readability are provided in Tables \ref{tbl:pyr}, \ref{tbl:resp}, and \ref{tbl:read}, respectively. The best correlation in each column has been specified in bold.

Analyzing the correlation results obtained by the best-performing AESOP metrics in Figure \ref{fig:pyr} show that \textsc{GRouge-2} achieves the best correlation with Pyramid, regarding the Spearman and Kendall rank correlations. However, \textsc{Rouge-su4} displays the best correlation with Pyramid considering the Pearson correlation. The key difference between the Pearson correlation and Spearman/Kendall rank correlation, is that the former assumes that the variables being tested are normally distributed, and linearly related to each other. The latter two measures are however non-parametric and make no assumptions about the distribution of the variables being tested. The assumption made by the Pearson correlation has been known too constraining \cite{ng2015better}, given that any two independent evaluation systems may not exhibit linearity.

Looking closer to the correlation with Pyramid scores, obtained by the variants of \textsc{Rouge}-based metrics in Table \ref{tbl:pyr}, we observe that every \textsc{GRouge} variant outperforms its corresponding \textsc{Rouge} and \textsc{Rouge-WE} variants, regardless of the correlation metric used. However, the only exception is \textsc{Rouge-su4}, which correlates slightly better with Pyramid when measuring with Pearson correlation. One possible reason is that Pyramid measures content similarity between peer and model summaries, while the variants of \textsc{GRouge} favor semantics behind the content for measuring similarities. Since some of the semantics attached to the skipped words are lost in the construction of skip-bigrams, \textsc{Rouge-su4} shows a better correlation comparing to \textsc{GRouge-su4}.

\begin{table}[h!]
	\centering
	\small
	\begin{tabular}{l*{14}{c}}
		\toprule
		\midrule
		\textbf{Metric} &\textbf{Pearson}&\textbf{Spearman}&  \textbf{Kendall}\\
		\midrule
		\textsc{Rouge-1} & 0.966 & 0.909 & 0.747 \\
		\textsc{Rouge-2} & 0.961 & 0.894 & 0.745 \\
		\textsc{Rouge-su4} & \textbf{0.981} & 0.894 & 0.737 \\
		\midrule
		\textsc{Rouge-WE-1} & 0.949 & 0.914 & 0.753 \\
		\textsc{Rouge-WE-2} & 0.977 & 0.898 & 0.744 \\
		\textsc{Rouge-WE-su4} & 0.978 & 0.881 & 0.720 \\
		\midrule
		\textsc{GRouge-1} & 0.968 & 0.916 & 0.758 \\
		\textsc{GRouge-2} & 0.979 & \textbf{0.921} & \textbf{0.768} \\
		\textsc{GRouge-su4} & 0.980 & 0.901 & 0.747 \\
		\midrule
		\bottomrule
	\end{tabular}
	\caption{Correlation of the \textbf{\textsc{Rouge}-based metrics} with the manual metric of Pyramid using the correlation metrics of Pearson $r$, Spearman $\rho$, and Kendall $\tau$ on the TAC 2011 AESOP dataset}
	\label{tbl:pyr}
\end{table}

Comparing the best-performing AESOP metrics in Figure \ref{fig:resp}, \textsc{GRouge-su4} achieves the best correlation with Responsiveness when measuring with the Pearson correlation. We also observe that \textsc{GRouge-2} obtains the best correlation with Responsiveness while measuring with the Spearman and Kendall rank correlations. The reason is that semantic interpretation of bigrams is easier, and that of contiguous bigrams is much more precise. Regarding Table \ref{tbl:resp}, every variant of \textsc{GRouge} outperforms its corresponding variant in the framework of \textsc{Rouge}.

\begin{table}[h!]
	\centering
	\small	
	\begin{tabular}{l*{14}{c}}
		\toprule
		\midrule
		\textbf{Metric} &\textbf{Pearson}&\textbf{Spearman}&  \textbf{Kendall}\\
		\midrule
		\textsc{Rouge-1} & 0.935 & 0.818 & 0.633 \\
		\textsc{Rouge-2} & 0.942 & 0.790 & 0.610 \\
		\textsc{Rouge-su4} & 0.955 & 0.790 & 0.602 \\
		\midrule
		\textsc{Rouge-WE-1} & 0.916 & 0.819 & 0.631 \\
		\textsc{Rouge-WE-2} & 0.953 & 0.797 & 0.615 \\
		\textsc{Rouge-WE-su4} & 0.954 & 0.787 & 0.597 \\
		\midrule
		\textsc{GRouge-1} & 0.940 & 0.822 & 0.635 \\
		\textsc{GRouge-2} & 0.954 & \textbf{0.863} & \textbf{0.705} \\
		\textsc{GRouge-su4} & \textbf{0.958} & 0.812 & 0.617 \\
		\midrule
		\bottomrule
	\end{tabular}
	\caption{Correlation of the \textbf{\textsc{Rouge}-based metrics} with the manual metric of Responsiveness using the correlation metrics of Pearson $r$, Spearman $\rho$, and Kendall $\tau$ on the TAC 2011 AESOP dataset}
	\label{tbl:resp}
\end{table}

The readability score reflects the fluency and structure of the summary, independently of content; and is based on grammaticality, structure, focus, coherence and etc.. According to Figure \ref{fig:read}, \textsc{GRouge-su4} and \textsc{GRouge-2} are superior to the best-performing AESOP metrics, regarding Pearson and Spearman/Kendall rank correlations, respectively. Although our main goal is not to improve the readability, \textsc{GRouge} achieves the best correlations with this metric. This is likely due to considering word types and part-of-speech tagging while aligning and disambiguating n-grams. Part-of-speech features are shown by \cite{feng2010comparison} to be helpful in the prediction of the linguistic quality. Measuring Readability in the \textsc{Rouge} framework, every variant of \textsc{GRouge} represents the best correlation results comparing to its corresponding variant of \textsc{Rouge} and \textsc{Rouge-WE} for all correlation metrics (Table \ref{tbl:read}).  

\begin{table}[h!]
	\centering	
	\small
	\begin{tabular}{l*{14}{c}}
		\toprule
		\midrule
		\textbf{Metric} &\textbf{Pearson}&\textbf{Spearman}&  \textbf{Kendall}\\
		\midrule
		\textsc{Rouge-1} & 0.790 & 0.391 & 0.285 \\
		\textsc{Rouge-2} & 0.752 & 0.398 & 0.293 \\
		\textsc{Rouge-su4} & 0.784 & 0.395 & 0.293 \\
		\midrule
		\textsc{Rouge-WE-1} & 0.785 & 0.431 & 0.322 \\
		\textsc{Rouge-WE-2} & 0.782 & 0.414 & 0.304 \\
		\textsc{Rouge-WE-su4} & 0.793 & 0.407 & 0.302 \\
		\midrule
		\textsc{GRouge-1} & 0.793 & 0.433 & 0.326 \\
		\textsc{GRouge-2} & 0.787 & \textbf{0.513} & \textbf{0.384} \\
		\textsc{GRouge-su4} & \textbf{0.824} & 0.440 & 0.334 \\
		\midrule
		\bottomrule
	\end{tabular}
	\caption{Correlation of the \textbf{\textsc{Rouge}-based metrics} with the manual metric of Readability using the correlation metrics of Pearson $r$, Spearman $\rho$, and Kendall $\tau$ on the TAC 2011 AESOP dataset}
	\label{tbl:read}
\end{table}

Overall, considering Pyramid, Responsiveness, and Readability, and regardless of the correlation metric used, every \textsc{GRouge} variant outperforms its corresponding \textsc{Rouge} variant, with only one exception: \textsc{Rouge-su4} correlates slightly better with Pyramid when measuring with Pearson correlation, to which possible reasons are discussed earlier. Looking at \textsc{GRouge-2} that is far more superior than its corresponding variants while measuring with Spearman and Kendall rank correlations, supports our proposal to consider semantics besides surface with \textsc{Rouge}. However, the large/small differences in competing correlations with human assessment are not an acceptable proof of superiority/inferiority in performance of one metric over another. Hence, prior to any conclusion in this regard, significance tests should be applied. 

\subsection{Significance Test}

Evaluation of summarization metrics depart from correlation with human judgment has included the ability of a metric/significance test combination to identify a significant difference between the quality of human and system-generated summaries \cite{rankel2011ranking}. To better clarify the effectiveness of \textsc{GRouge}, we use pairwise Williams significance test\footnote{Also known as Hotelling-Williams} recommended by \cite{graham2015re} for summarization evaluation. Accordingly, evaluation of a given summarization metric, $M_{new}$, takes the form of quantifying three correlations: $r(M_{new}, H)$, that exists between the evaluation metric scores for summarization systems and corresponding human assessment scores; $r(M_{base}, H)$, that stands for the correlation of baseline metrics with human judges; and the third correlation, between evaluation metric scores themselves, $r(M_{base}, M_{new})$. It can happen for a pair of competing metrics for which the correlation between metric scores is strong, that a small difference in competing correlations with human assessment is significant, while, for a different pair of metrics with a larger difference in correlation, the difference is not significant \cite{graham2015re}. Utilizing this significance test, the results show that all increases in correlations of \textsc{GRouge} compared to \textsc{Rouge} and \textsc{Rouge-WE} variants in Tables \ref{tbl:pyr}, \ref{tbl:resp} and \ref{tbl:read} are statistically significant ($p < 0.05$).

\subsection{Exploring Scaling Factor}\label{sec:explor}

\begin{figure}[h!]
	\centering
	\includegraphics[scale=0.41]{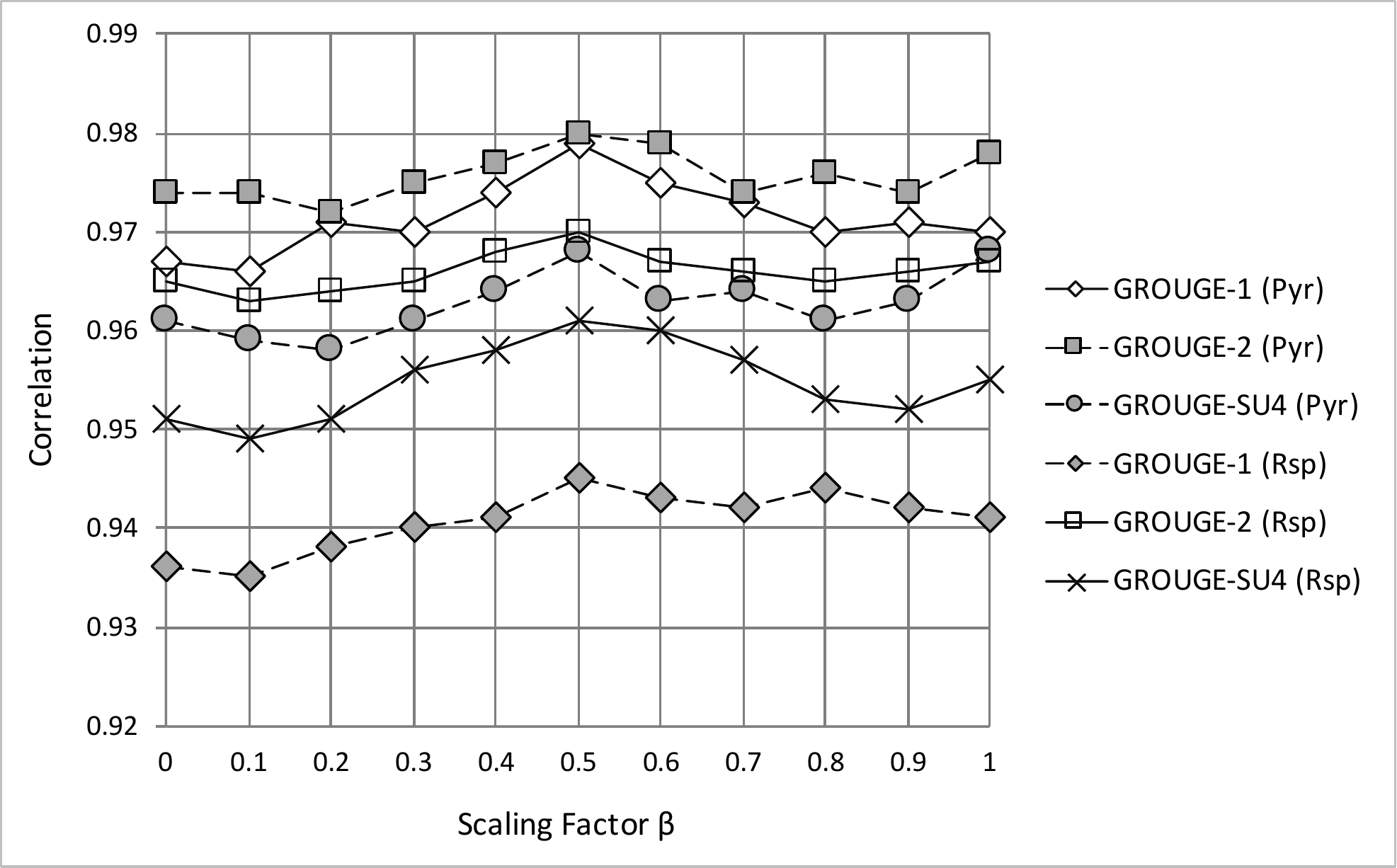}
	\caption{Exploring scaling factor $\beta$ on the TAC 2010 AESOP dataset}
	\label{fig:tunning}
\end{figure}

In this section, we optimize scaling factor $\beta$ in Equation \ref{eq:finalsim}, and obtain a balance between contributions of lexical and semantic similarity scores to calculate the lexico-semantic similarity. To this end, we make use of the TAC 2010 AESOP dataset. Figure \ref{fig:tunning} shows the correlation results obtained by the variants of \textsc{GRouge} with Pyramid (Pyr) and Responsiveness (Rsp) metrics measured by Pearson. The best results are observed when using $\beta = 0.5$. Performance deteriorates when the value of $\beta$ approaches 1.0 which indicates the \textsc{Rouge} scores without any touch of semantic similarity. Decreasing the weight of $\beta$ to zero causes the exclusion of lexical match counts, and consequently inappropriateness of the outcomes. This demonstrates the importance of using both lexical and semantic similarities to fairly judge the quality of summaries.

\section{Conclusion}\label{sec:concl}

We have proposed an effective approach (namely \textsc{GRouge}) to overcome the limitation of high lexical dependency in \textsc{Rouge}. We improve on \textsc{Rouge} by performing both semantic and lexical analysis of summaries. Evaluation is processed by comparing each model-gram against the corresponding peer summary text. To this end, the PPR algorithm is employed, and all senses have been disambiguated before comparison. Experiments over the TAC AESOP datasets demonstrate that \textsc{GRouge} achieves higher correlations with manual judgments in comparison with the well-established \textsc{Rouge}. Since this approach goes beyond the lexical surface and exploits the underlying semantics, we believe that it would work even better on more comprehensive texts such as a dataset provided for the evaluation of abstractive summaries. Therefore, our ongoing work includes constructing a standard dataset for assessing the automatic metrics specified to evaluate abstractive summaries. We also believe that this approach can open a door to the evaluation of automatic text simplification. The reason is that text simplification indicates the process of simplifying a text without losing its meaning, and this approach can capture the underlying meaning in a text, regardless of its surface. Hence, in future, we intend to adopt this approach with the aim of helping \textsc{Rouge} to gain qualitative insights into the nature of text simplification.  

\bibliographystyle{plain} 
\bibliography{sample}

\end{document}